%% file: main_arxiv.tex
\documentclass[10pt,twocolumn,letterpaper]{article}

\usepackage[letterpaper,margin=0.75in]{geometry}
\usepackage[T1]{fontenc}
\usepackage{mathptmx}
\usepackage{graphicx}
\usepackage{amsmath,amssymb}
\usepackage{booktabs}
\usepackage{multirow}
\usepackage{stfloats}
\usepackage{placeins}
\usepackage{float}
\usepackage{flafter}
\usepackage{url}
\usepackage{xcolor}
\usepackage[numbers,sort&compress]{natbib}
\usepackage[breaklinks,colorlinks,allcolors=blue]{hyperref}

\input{math_commands.tex}

\newcommand{\xref}[1]{\ref{#1}}

\renewenvironment{abstract}{%
  \begin{center}\bfseries\abstractname\end{center}%
  \begin{list}{}{\setlength{\leftmargin}{1.1em}\setlength{\rightmargin}{1.1em}}\item\relax
}{%
  \end{list}}
\newcommand{\runinhead}[1]{\par\addvspace{1.0ex}\noindent\textbf{#1}\enspace\nobreak}

\usepackage{titlesec}
\titleformat{\section}{\normalfont\large\bfseries}{\thesection}{0.6em}{}
\titlespacing*{\section}{0pt}{1.6ex plus .3ex minus .2ex}{0.8ex plus .1ex}
\titleformat{\subsection}{\normalfont\normalsize\bfseries}{\thesubsection}{0.5em}{}
\titlespacing*{\subsection}{0pt}{1.2ex plus .2ex minus .1ex}{0.5ex}

\def\WITHAPPENDIX{}
\begin{document}

\twocolumn[{%
\rule{\textwidth}{0.5pt}
\vspace{0.85em}
\begin{center}
{\LARGE\bfseries ACiT: Proprioceptive-Prior Action Chunking with Inverted Transformers\par}
\vspace{1.15em}
{\large Haixiang Wu\textsuperscript{1}\quad Zhihao Guan\textsuperscript{1}\quad Yang Yang\textsuperscript{1}\par}
\vspace{0.55em}
{\small \textsuperscript{1}Nanjing University of Science and Technology, Nanjing, China\par}
\end{center}
\vspace{0.35em}
\rule{\textwidth}{0.5pt}
\vspace{0.9em}
}]

\begin{abstract}
Visuomotor policies condition action prediction on visual and proprioceptive observations. While historical proprioceptive sequences carry rich dynamic priors, current methods treat them as equal to high-dimensional vision, causing the proprioceptive signal to be overwhelmed. To identify what is lost in equal-conditioning fusion, we analyze proprioceptive data and find that strong inter-joint coupling is the key action-determining structure. This finding aligns with the neural sensorimotor loop, in which proprioception first forms a body-centered prior and vision is later calibrated via parietal remapping. Inspired by this, we propose \textbf{ACiT}, Proprioceptive-Prior Action Chunking with Inverted Transformers, which reinforces historical joint-coupling as the primary condition with vision as a later calibration. This design is instantiated through three modules: Inverted State Encoding builds a proprioceptive prior via per-joint tokens and inter-joint self-attention. State-Visual Grounding lets joint tokens query visual evidence, making vision goal- and calibration-only. Inverted Action Chunking decodes per-joint trajectories independently. ACiT improves success rates on most LIBERO, RoboTwin 2.0, and real-world tasks, with the largest gains on those that require dynamics and coordination. The same conditioning design transferred to the VLA backbone $\pi_0$-ACiT yields significant gains with better execution efficiency.
\end{abstract}
\vspace{0.2em}
\setcounter{dbltopnumber}{5}
\renewcommand{\dbltopfraction}{1.0}
\setcounter{topnumber}{3}
\raggedbottom

\begin{figure*}[t]
	\begin{center}
		\includegraphics[width=\textwidth]{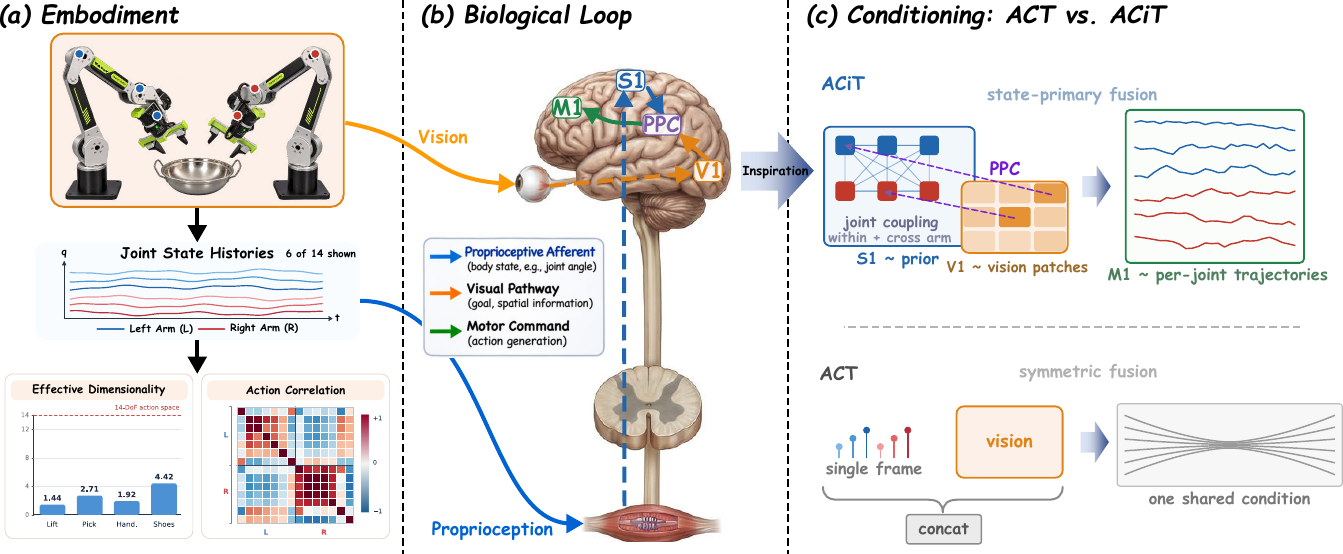}
	\end{center}
\caption{Motivation. \textbf{(a) Embodiment}: a dual-arm robot has 14 joints, each with its own state history, while the action space has a much lower effective dimensionality (PR 1.44--4.42) and strong inter-joint coupling, the state therefore should not be compressed into one shared vector. \textbf{(b) Biological inspiration}: proprioception forms a body-centered prior in S1, vision reaches PPC via V1 only to remap and calibrate the goal, and M1 issues the command. The two modalities are not symmetric, and their order is fixed. \textbf{(c) Conditioning interface}: ACT concatenates a single-frame state with vision into one condition shared by all joints; ACiT keeps the per-joint prior as the primary condition, retrieves visual evidence per joint, and decodes one trajectory per joint.}
\label{fig:intro}
\end{figure*}
\section{Introduction}
Visuomotor policies predict actions from multi-view observations and
proprioception. Early work conditioned on a single current state, where ACT maps
the current joint state and images to an action chunk~\citep{act}. Diffusion
Policy conditions on a short observation history instead~\citep{dp}. A single frame proved insufficient to resolve action ambiguity, while
history carries dynamics and hidden-state information, so proprioceptive history
was explicitly incorporated. Recent models such as A2A and Chronos explicitly model proprioceptive history~\citep{a2a,chronos}.
Another line discretizes historical states or actions into tokens to better
integrate with LLM backbones, at the cost of losing continuous coupling
structure~\citep{fast,vita,vqbet}. Reconstruction-only tokenization can also
erode control-relevant information~\citep{salt}. These works validate the value
of using history from both continuous and discrete directions, and state history
is becoming a consensus for action prediction.

These advances have not converged at the conditioning interface. First,
mainstream methods map historical proprioceptive states as an independent
modality, either discretely or continuously, and place them alongside a pooled
global visual vector as conditions for action generation. This naive fusion lets
high-dimensional, low-SNR visual features entangle and dominate the conditioning,
leading to degraded robustness or spurious
shortcuts~\citep{slotflow,imprint}. Second, we analyze the state history and the
action itself: the effective dimensionality of the action distribution is far
below the number of joints, and joints are strongly coupled,
as shown in Fig.~\ref{fig:intro}\((a)\). However, existing
methods model temporal dependencies and pack all joints into one vector, so the
inter-joint coupling, a structure critical to action, can only be implicit
inside the token.

Therefore, how to encode historical state information, how to fuse modalities,
and how to feed the result into action decoding is an open problem. Inspired by
neuroscience, the human chain from peripheral proprioception to body-centered
action generation is as follows (Fig.~\ref{fig:intro}\((b)\)): muscle and joint receptors
continuously encode joint angle, velocity, and tension and project through
large-diameter myelinated afferents~\citep{proske2012proprioceptors,windhorst2007muscle}.
The somatosensory cortex and posterior parietal cortex aggregate proprioceptive
history into a body-state representation, forming a body-centered
prior~\citep{avillac2005reference,asada2018proprioception}. The posterior parietal
cortex then remaps visual goals from retinal coordinates into
body-centered coordinates, with proprioceptive context modulating this
remapping~\citep{avillac2005reference,heed2015tactile}. Finally, the
parietal-motor pathway sends the calibrated body-centered representation to
action generation~\citep{wolpert1995internal,miall1996forward}. The key
computational asymmetry is that proprioception provides a continuous,
low-latency body-state stream, whereas visual goals are remapped and integrated
into the body-centered representation afterward.

We map this sensorimotor loop to Bayesian cue integration: the fast, continuous
proprioceptive stream forms a prior over the body state, visual goals provide
goal evidence (a likelihood-like term), and calibrating the prior yields a
posterior-like body-centered representation for action generation. Based on
this, we propose ACiT on top of ACT, the smallest architecture that still
carries the full interface (multi-view images, proprioception, and a chunked
action decoder), so that the encoder and the head stay fixed while only the
conditioning changes. It has three modules. First, Inverted State
Encoding compresses each joint's entire history into a token and directly
models inter-joint coupling with self-attention. Second, State-Visual
Grounding treats vision as a per-joint goal-calibration signal and does not let
it directly enter the encoder. Third, Inverted Action Chunking treats
each joint as a token on the action side, and each joint independently generates
its complete trajectory chunk. The decoder does not need to translate between
``visual coordinates'' and ``joint coordinates''. The historical proprioceptive
sequence and action generation share a more consistent feature space.
To separate the interface from the backbone, we port the same conditioning to a
pretrained VLA and obtain $\pi_0$-ACiT, which keeps $\pi_0$'s architecture and changes
only how the state enters it.
Experiments on LIBERO, RoboTwin~2.0, and real-world manipulation tasks show that
ACiT and $\pi_0$-ACiT improve success on most tasks, with improved robustness
under visual perturbations, especially on tasks that depend on dynamics and
joint coordination.

Our contributions are as follows:

\begin{itemize}
\item We formulate the conditioning interface as Bayesian sensorimotor cue integration, with history as the prior and vision as the calibration evidence, and realize it as an explicit asymmetric fusion.
\item We propose ACiT, which uses three inverted Transformer modules to
explicitly build and preserve the inter-joint coupling dynamics in the state
history instead of flattening them.
\item We validate ACiT on LIBERO, RoboTwin~2.0, and real-world tasks, where it
improves the average success rate by 5.7 to 10.8 points over the strongest
baseline. $\pi_0$-ACiT improves the average success rate by 26.9 points at
20.2\% lower inference latency than $\pi_0$.
\end{itemize}

\section{Related Work}
\label{sec:related}
\runinhead{History in the conditioning interface.}%
Early visuomotor policies condition
on a single frame: ACT maps the current joint state and images to an action
chunk~\citep{act}, and Diffusion Policy instead uses a short observation
window~\citep{dp}. Later work makes the state history explicit in the conditioning
input~\citep{a2a,chronos}. A parallel line discretizes states or
actions so that autoregressive backbones can consume them~\citep{vqbet,fast,omnisat,vita},
at the cost of losing continuous structure, and reconstruction-only tokenization can
erode control-relevant information~\citep{salt}. Under both interfaces proprioception
enters as one flat vector or a one-dimensional token sequence, so the coupling among
joints stays implicit.

\runinhead{Vision and proprioception under fusion.}%
Pooling both modalities into one
condition leaves nothing to stop the high-dimensional visual stream from dominating
and introducing spurious shortcuts~\citep{slotflow,imprint}. One line responds by
letting the robot state steer vision, through robot-centric pooling~\citep{rcp} or
active visual attention~\citep{avavla}, yet the asymmetry is applied to perception
rather than to the conditioning itself. Large VLA backbones inherit the same
symmetric interface, concatenating one proprioceptive token with visual and language
tokens~\citep{pi0,pi05,groot}.

\runinhead{Structure as an inductive bias.}%
Topology-aware models inject the kinematic
graph as a structural bias for cross-embodiment
transfer~\citep{getzero,human2locoman}. ACiT shares that motivation but acts on the
conditioning interface: it carries history per joint, keeps the resulting prior
primary, and lets each joint retrieve its own visual calibration in a single
generation pipeline. Appendix~\xref{app:related} expands the comparison.

\section{Method}

\begin{figure*}[t]
\centering
\includegraphics[width=\textwidth]{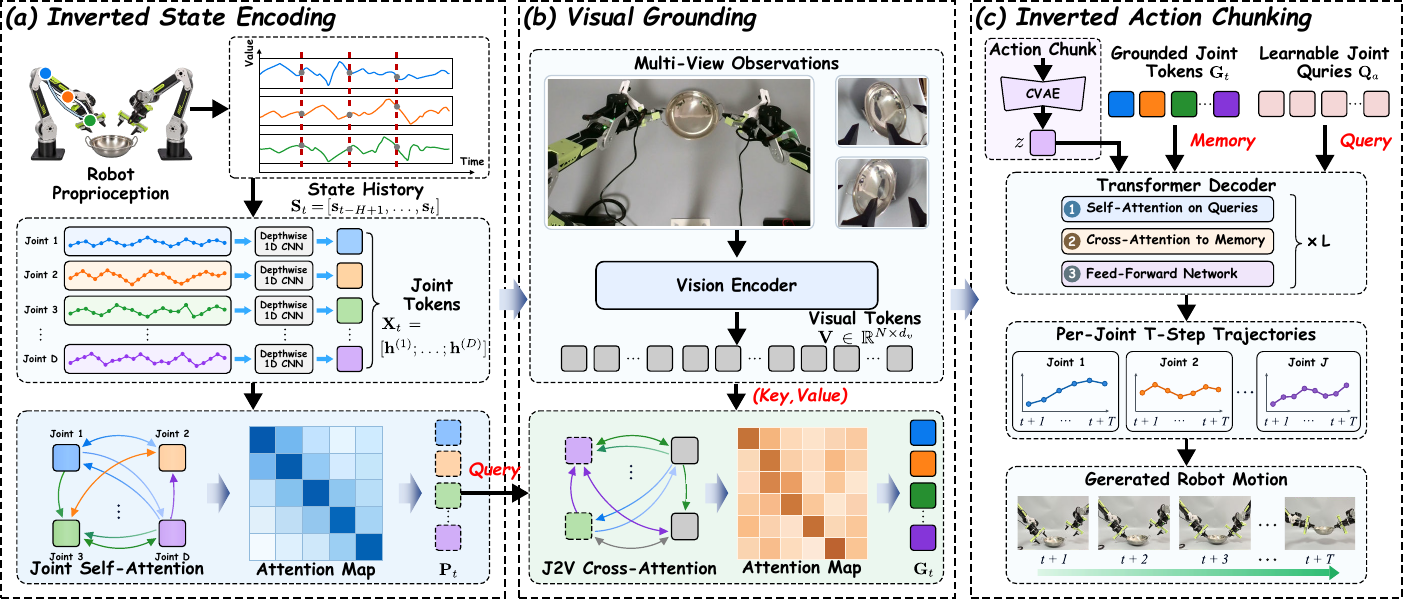}
\caption{ACiT architecture, organized by its three modules. \textbf{(a) Inverted State Encoding}: per-joint depthwise 1D convolutions turn each joint's history into one token, and self-attention among the tokens gives the proprioceptive prior $\mathbf{P}_t$. \textbf{(b) State-Visual Grounding}: the joint tokens query the visual tokens of the multi-view observations and produce grounded tokens $\mathbf{G}_t$. \textbf{(c) Inverted Action Chunking}: $\mathbf{G}_t$ is the memory of a transformer decoder whose learnable joint queries output one $T$-step trajectory per joint.}
\label{fig:method}
\end{figure*}

\subsection{Preliminaries}

Let the robot have $D$ joints, with state $\mathbf{s}_t \in \mathbb{R}^{D}$ and
action $\mathbf{a}_t \in \mathbb{R}^{D}$ at time $t$. An action-chunking policy
predicts
\begin{equation}
\mathbf{A}_t = [\mathbf{a}_t, \dots, \mathbf{a}_{t+T-1}] \in \mathbb{R}^{T \times D},
\end{equation}
and executes its first few steps. The state history is
$\mathbf{S}_t = [\mathbf{s}_{t-H+1}, \dots, \mathbf{s}_t] \in \mathbb{R}^{H \times D}$, where $H$ is the history length (number of past frames). $\mathbf{V} \in \mathbb{R}^{N \times d_v}$ denotes the visual token matrix, whose rows are the $N$ visual tokens.

Most chunk-based policies share one skeleton~\citep{act,dp}: a Transformer
encoder~\citep{vaswani} maps the observations $\mathbf{O}_t$ to a condition, and a
decoder emits the chunk,
$\,p(\mathbf{A}_t \mid \mathbf{O}_t) = \mathrm{Dec}_\theta\bigl(\mathrm{Enc}_\phi(\mathbf{O}_t)\bigr)$.
What differs across methods is how the proprioceptive history is represented
inside that condition, which is the axis ACiT changes. The encoder and the
decoder are those of the host policy (Sec.~\xref{sec:related-proprio}). The
equal-conditioning baseline we depart from places the current state and vision
side by side as conditions:
\begin{equation}
\pi_\theta\bigl(\mathbf{A}_t \mid \mathbf{s}_t,\ \mathbf{V}\bigr).
\label{eq:baseline}
\end{equation}

\runinhead{Bayesian Sensorimotor Formulation.}%
Let $\mathbf{b}_t$ be the latent
body-centered state and $g$ the task goal (vision, or vision and language in a
VLA). The ideal visuomotor computation factorizes into two stages,
\begin{align}
p(\mathbf{b}_t \mid \mathbf{S}_t,\mathbf{V})
&\propto p(\mathbf{b}_t \mid \mathbf{S}_t)\,
p(\mathbf{V} \mid \mathbf{b}_t,g), \label{eq:bayes-posterior} \\
p(\mathbf{A}_t \mid \mathbf{S}_t,\mathbf{V})
&= \int p(\mathbf{A}_t \mid \mathbf{b}_t)\,
p(\mathbf{b}_t \mid \mathbf{S}_t,\mathbf{V})\,d\mathbf{b}_t.
\label{eq:bayes-policy}
\end{align}
Equation~\eqref{eq:bayes-posterior} states that visual evidence calibrates a
proprioceptive prior into a posterior body-state estimate, and
Equation~\eqref{eq:bayes-policy} marginalizes over this posterior to produce the
action distribution. This formulation mirrors Bayesian cue integration in
sensorimotor control~\citep{vanbeers1999integration,kording2004bayesian}.

\runinhead{ACiT Instantiation.}%
Rather than modeling explicit probability
distributions, ACiT uses deterministic representations to instantiate this
computation:
\begin{equation}
\pi_\theta\bigl(\mathbf{A}_t \mid \mathbf{G}_t\bigr), \qquad
\mathbf{G}_t = f_\theta(\mathbf{S}_t,\ \mathbf{V}),
\label{eq:acit}
\end{equation}
where $\mathbf{G}_t$ is a deterministic representation of the posterior-like
body state in Eq.~\eqref{eq:bayes-posterior}, and the action head
$\pi_\theta(\mathbf{A}_t \mid \mathbf{G}_t)$ replaces the integration in
Eq.~\eqref{eq:bayes-policy}. The function $f_\theta$ is factorized into
per-joint history encoding with a body-centered prior
(Sec.~\ref{sec:state-encoding}) followed by state-visual grounding
(Sec.~\ref{sec:grounding}). The action head consumes
$\mathbf{G}_t$ in Sec.~\ref{sec:action-chunking}. This is a computational
analogy, not a literal probabilistic model. $\mathbf{b}_t$ is distinct from the
CVAE latent $z$ used for training.

\subsection{Inverted State Encoding}
\label{sec:state-encoding}

For each joint $j$, ACiT applies a depthwise 1D convolution along time with
kernel size $\kappa$, followed by a shared temporal projection:
\begin{equation}
u_i^{(j)} = \sum_{\tau=-\lfloor \kappa/2 \rfloor}^{\lfloor \kappa/2 \rfloor}
k_{\tau}^{(j)}\, \bar{s}_{i+\tau}^{(j)},
\qquad i = 1,\dots,H,
\end{equation}
where $\kappa$ is the kernel size, $\bar{s}_i^{(j)} = s_{t-H+i}^{(j)}$ is the
$i$-th frame in the history window, $k_{\tau}^{(j)}$ is the joint-specific
convolution weight at offset $\tau$, and zero padding is applied at both ends.
We use $\kappa=3$ by default, giving weights
$\mathbf{k}^{(j)} = [k_{-1}^{(j)}, k_0^{(j)}, k_1^{(j)}]$. The convolved features
are stacked into
$\mathbf{u}^{(j)} = [u_1^{(j)},\dots,u_H^{(j)}]^\top \in \mathbb{R}^{H}$.

A shared linear layer and a joint-identity embedding then project the temporal
feature to a $d$-dimensional token:
\begin{equation}
\mathbf{h}^{(j)} = W_{\mathrm{time}}\,\mathbf{u}^{(j)} + \mathbf{e}^{(j)}
\in \mathbb{R}^{d},
\end{equation}
where $W_{\mathrm{time}} \in \mathbb{R}^{d \times H}$ is shared across joints and
$\mathbf{e}^{(j)} \in \mathbb{R}^{d}$ is a learnable joint-identity embedding.
Stacking all joint tokens gives
$\mathbf{X}_t = [\mathbf{h}^{(1)};\dots;\mathbf{h}^{(D)}] \in \mathbb{R}^{D \times d}$.
The depthwise convolution captures local temporal patterns such as velocity and
acceleration without mixing joints. The history is in effect a multivariate time
series with one channel per joint, and the coupling we care about is a dependency
between channels rather than between time steps. Treating each channel as a token
and letting attention operate across channels is the inverted formulation that
has been shown to capture variate-wise structure in forecasting~\citep{itransformer}. Here the variates are joints, and their dependency is the coupling that determines
the action.

\runinhead{Proprioceptive prior.}%
The joint tokens then attend to each other, so that the coupling structure is
represented explicitly rather than hidden inside a flattened vector:
\begin{align}
\mathbf{Q} &= \mathbf{X}_t W_Q, \quad
\mathbf{K} = \mathbf{X}_t W_K, \quad
\mathbf{Z}_x = \mathbf{X}_t W_V, \\
\mathbf{P}_t &= \mathrm{softmax}\!\left(
\frac{\mathbf{Q}\mathbf{K}^{\top}}{\sqrt{d_k}}\right)\mathbf{Z}_x
\in \mathbb{R}^{D \times d},
\end{align}
where $\mathbf{Q},\mathbf{K},\mathbf{Z}_x$ are the queries, keys and values, $d_k$ is the key
dimension, and $\mathbf{P}_t = [\tilde{\mathbf{h}}^{(1)};\dots;\tilde{\mathbf{h}}^{(D)}]$
is a deterministic representation of the prior $p(\mathbf{b}_t \mid \mathbf{S}_t)$
in Eq.~\eqref{eq:bayes-posterior}, i.e., the body-centered prior from which the
action is eventually generated.

\subsection{State-Visual Grounding}
\label{sec:grounding}

Given the prior $\mathbf{P}_t$, vision now enters only as evidence. Each prior
token serves as a query to the visual tokens:
\begin{align}
\mathbf{Q}_c &= \mathbf{P}_t W_Q^{c}, \quad
\mathbf{K}_c = \mathbf{V} W_K^{c}, \quad
\mathbf{Z}_c = \mathbf{V} W_V^{c}, \\
\mathbf{G}_t &= \mathrm{softmax}\!\left(
\frac{\mathbf{Q}_c\mathbf{K}_c^{\top}}{\sqrt{d_k}}\right)\mathbf{Z}_c
\in \mathbb{R}^{D \times d}. \label{eq:grounding}
\end{align}
Equation~\eqref{eq:grounding} is the concrete instantiation of the mapping
$f_\theta$ in Eq.~\eqref{eq:acit}, where
$\mathbf{G}_t = [\mathbf{c}^{(1)};\dots;\mathbf{c}^{(D)}]$ is the grounded
condition and a deterministic representation of the posterior-like body state in
Eq.~\eqref{eq:bayes-posterior}. All blocks use pre-LN residual connections followed by a feed-forward network.
Visual evidence is thus retrieved into
a body-centered state rather than fused symmetrically with proprioception. This
is consistent with visual targets being remapped into body-centered reference
frames~\citep{avillac2005reference,heed2015tactile} and with vision modulating,
rather than replacing, the body schema~\citep{asada2018proprioception}.

\subsection{Inverted Action Chunking}
\label{sec:action-chunking}

The action head is interchangeable. With the ACT transformer decoder, $D$
learned joint queries $\mathbf{Q}_a \in \mathbb{R}^{D \times d}$ attend to
the memory $\mathbf{M} = [\mathbf{G}_t;\mathbf{m}]$. As in the self- and
cross-attention blocks above, each head computes
\begin{equation}
\mathbf{O} = \mathrm{softmax}\!\left(
\frac{\mathbf{Q}_a \mathbf{K}_m^{\top}}{\sqrt{d_k}}\right)\mathbf{Z}_m
\in \mathbb{R}^{D \times d},
\end{equation}
where $\mathbf{K}_m = \mathbf{M}W_K^{a}$,
$\mathbf{Z}_m = \mathbf{M}W_V^{a}$, and
$\mathbf{m} = W_m z$ is a learned token embedding of the CVAE latent $z$
defined below. The CVAE latent is used only here. State-visual grounding is
independent of $z$. With a conditional 1D U-Net, the denoiser is conditioned
on $\mathbf{G}_t$ and $\mathbf{m}$ and produces the same per-joint output layout. In both cases, a shared linear
projection maps every joint output to a full trajectory:
\begin{equation}
\hat{\mathbf{a}}^{(j)} = W_o\, \mathbf{o}^{(j)},
\qquad W_o \in \mathbb{R}^{T \times d},
\end{equation}
and concatenation gives
$\hat{\mathbf{A}}_t = [\hat{\mathbf{a}}^{(1)};\dots;\hat{\mathbf{a}}^{(D)}] \in
\mathbb{R}^{T \times D}$. The action is thus modeled as $D$ joint trajectories
rather than $T$ time-step vectors.

\subsection{Training Objective}

Following ACT~\citep{act}, we use a CVAE~\citep{sohn2015learning} latent
$z \sim q_\phi(z \mid \mathbf{S}_t,\mathbf{A}_t)
= \mathcal{N}(\boldsymbol{\mu}_\phi,\,
\mathrm{diag}(\boldsymbol{\sigma}_\phi^{2}))$,
where $z = \boldsymbol{\mu}_\phi + \boldsymbol{\sigma}_\phi \odot
\boldsymbol{\epsilon}$ with
$\boldsymbol{\epsilon} \sim \mathcal{N}(\mathbf{0},\mathbf{I})$, and
$z=\mathbf{0}$ at inference. The objective is the standard CVAE
loss~\citep{sohn2015learning}:
\begin{equation}
\mathcal{L} = \mathbb{E}_{\mathcal{D}}\Big[
\tfrac{1}{TD}\|\hat{\mathbf{A}}_t-\mathbf{A}_t\|_1
+ \lambda\,\mathrm{KL}(q_\phi\|p)\Big],
\end{equation}
where $\mathbb{E}_{\mathcal{D}}$ is the expectation over demonstrations,
$q_\phi$ abbreviates $q_\phi(z \mid \mathbf{S}_t,\mathbf{A}_t)$,
$p$ abbreviates $p(z)=\mathcal{N}(\mathbf{0},\mathbf{I})$,
$\mathrm{KL}$ denotes the Kullback-Leibler divergence, and $\lambda$ sets the
weight of this term.

\section{Experiments}
\label{sec:exp}

\subsection{Experimental Setup}

\begin{figure}[t]
\centering
\includegraphics[width=\columnwidth, trim=0 14 0 0, clip]{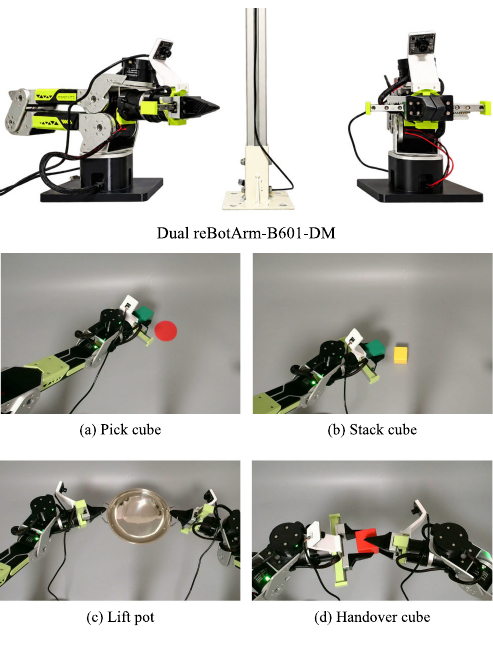}
\caption{Real-world platform and tasks. The \textbf{Dual reBotArm-B601-DM} performs four tasks: (a) pick cube, (b) stack cube, (c) lift pot, and (d) handover cube.}
\label{fig:real-hardware}
\end{figure}

\begin{figure}[t]
\centering
\includegraphics[width=\columnwidth]{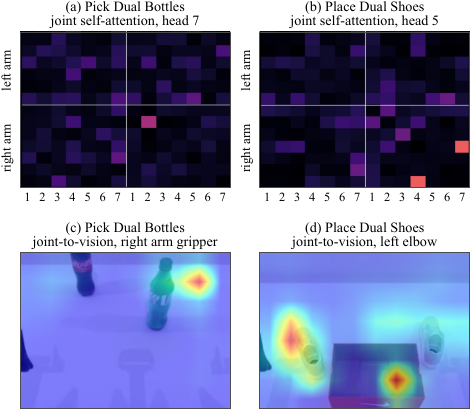}
\caption{Joint attention of ACiT. Top: joint self-attention inside the Inverted
State Encoding on two tasks, averaged over five episodes and ten frames, with the
left arm (joints 1--7) and the right arm (joints 8--14) marked. Bottom:
joint-to-vision cross attention inside State-Visual Grounding, for the gripper of
the right arm and for the elbow of the left arm.}
\label{fig:attn}
\end{figure}

\runinhead{Benchmarks and Tasks.}%
We evaluate on two simulation benchmarks and a real-world dual-arm platform.
\textbf{RoboTwin~2.0}~\citep{robotwin2} is a dual-arm manipulation benchmark with
14-DoF joint states and actions and three cameras (a top camera and two wrist
cameras). We use six tasks: Lift Pot, Pick Dual Bottles, Handover Block,
Move Can Pot, Place Dual Shoes and Shake Bottle, abbreviated as Lift, Pick Dual,
Handover, Move Can, Shoes and Shake. Each task has an easy and a randomized hard
split, the latter varying object pose, color and background (Appendix~\xref{app:sim-tasks}).
\textbf{LIBERO}~\citep{libero} is a single-arm benchmark with 7-DoF actions,
8-DoF states, and two cameras. We evaluate on Spatial, Object, and LIBERO-10.
We exclude Goal, which requires language instructions.

\runinhead{Real-World Platform.}%
We build a dual-arm platform from two reBot Arm
B601-DM arms~\citep{rebot}. Each arm has six DoF, a 1.5\,kg payload, and
$\pm0.2$\,mm repeatability. A global camera and two wrist cameras provide
observations. We evaluate four tasks: pick cube and stack cube with one arm, and
lift pot and handover cube with two arms. Hyperparameters and per-task
protocols are in Appendix~\xref{app:impl}. 

\runinhead{Baselines and Comparisons.}%
For small action policies we compare three baselines.
ACT~\citep{act} uses a CVAE and a transformer decoder on the ACT backbone and
conditions on the current frame only. Diffusion Policy~\citep{dp} and the
flow-matching baseline~\citep{flowmatching} share the same observation encoder
and the same 1-D U-Net, and both condition on a proprioceptive history. They
differ only in the generative objective (DDPM~\citep{ddpm} versus flow matching~\citep{flowmatching})
and in the sampler. The task-specific policies are trained from scratch with LeRobot~\citep{lerobot} in
PyTorch~\citep{pytorch}, on a shared ResNet-18~\citep{resnet} backbone and with
AdamW~\citep{adamw}. The VLA keeps its PaliGemma backbone. Simulation runs in
MuJoCo~\citep{mujoco} through RoboTwin~\citep{robotwin2}.

\subsection{Main Results on Simulation Benchmarks}
\label{sec:main}

\begin{figure*}[t]
\centering
\includegraphics[width=\textwidth]{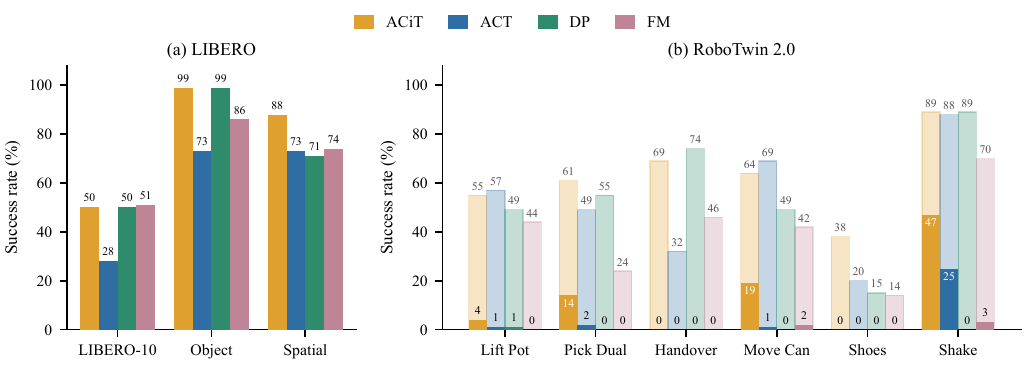}
\caption{Main results on the simulation benchmarks. Left: LIBERO. Right: RoboTwin~2.0. Pale bars denote easy and solid bars denote randomized hard. A number above a pale bar is its easy rate, and a number on a solid bar is its hard rate. ACiT is first in each group.}
\label{fig:main-results}
\end{figure*}

Figure~\ref{fig:main-results} compares ACiT with ACT, DP, and FM on LIBERO and
RoboTwin~2.0. On LIBERO the gain appears on Spatial, where success depends on
precise placement and therefore needs a different visual reference for each
joint. On Object the strong baselines are near saturation and ACiT matches them, while on
LIBERO-10 the four methods are close and the LIBERO mean is carried by Spatial.
On RoboTwin~2.0 the pattern is sharper: ACiT leads on the tasks limited by
inter-arm coordination (Pick Dual Bottles, Place Dual Shoes), and its margin over the
strongest baseline grows under randomization on the tasks limited by velocity and
phase (Shake Bottle, Move Can Pot), where a flat condition cannot express the motion
that the scene excites.

\runinhead{What the Attention Does.}%
Figure~\ref{fig:attn} shows what the two attention mechanisms do. In the top row the
joint self-attention is sparse: a few pairs carry most of it, and they are not limited
to neighboring joints, including pairs that span the two arms.
Joint-level tokens carry this structure, which a flattened state vector cannot. In the
bottom row a joint's query does not spread over the frame but lands on the objects
its motion depends on, the bottle a gripper takes in Pick Dual Bottles and, in Place
Dual Shoes, both the shoe to lift and the box to place it in. Vision enters as
evidence addressed to a joint, not as one summary of the scene.

\begin{figure*}[t]
\centering
\includegraphics[width=\textwidth]{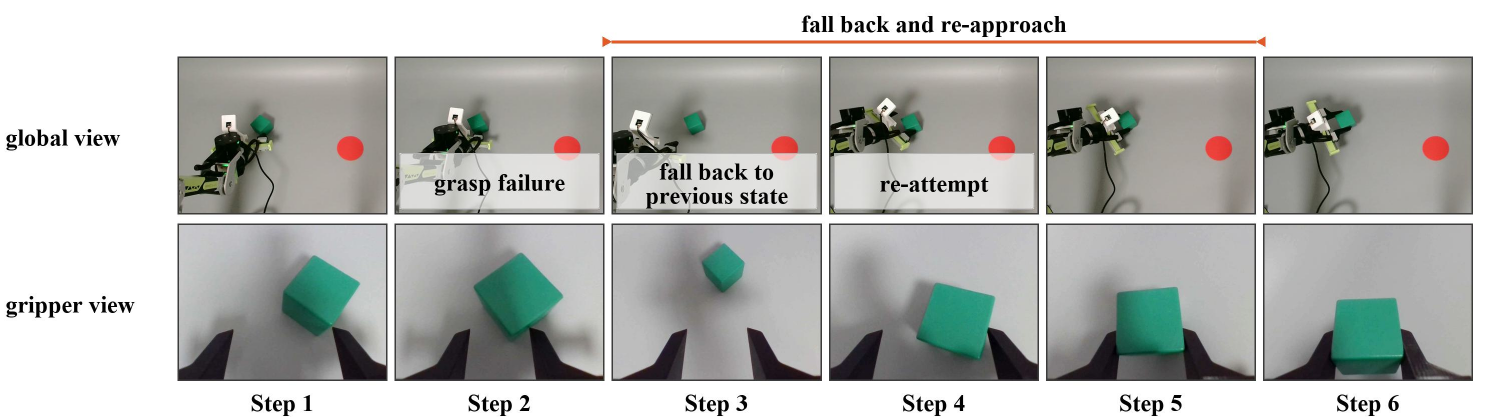}
\caption{Closed-loop recovery on the real robot. Top: global view. Bottom: gripper view. After a grasp failure (Step 2), ACiT falls back to the previous state (Step 3) and re-attempts (Step 4), completing the pick (Step 6).}
\label{fig:real-search}
\end{figure*}

\subsection{Real-World Experiments}
\label{sec:real-world}

Fig.~\ref{fig:real-hardware} shows the four real-world tasks, and one of the most direct pieces of
evidence of the gain appears in failure recovery. Figure~\ref{fig:real-search} is a
closed-loop trace of ACiT: the grasp fails at Step 2, the arm falls back to the
previous state at Step 3 and re-attempts at Step 4, completing the pick at
Step 6. In the rollouts we inspected, the baselines do not back off at the
failure point and instead keep searching in place.

A failed grasp is a proprioceptive event: the gripper closes to its limit
without an object and the arm does not take the load it expects, while vision
stays misleading because the cube is still in the workspace, only displaced.
Under equal conditioning the visual evidence can dominate the proprioceptive one,
which leaves the policy holding the pose or closing again. ACiT lets
proprioception carry the state and reduces vision to a per-joint calibration, so
the condition reflects a closed and empty gripper, and the decoded chunk is the
back-off-and-retry segment that the demonstrations contain.

\begin{figure*}[t]
\centering
\includegraphics[width=\textwidth]{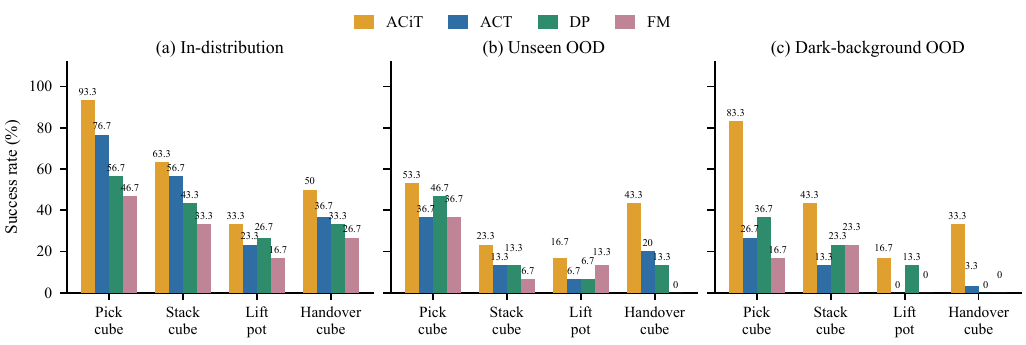}
\caption{Real-world results. Success rates for ACiT, ACT, DP, and FM on four tasks under in-distribution, unseen-OOD, and dark-background-OOD conditions. ACiT is shown first in each group.}
\label{fig:real-results}
\end{figure*}

Figure~\ref{fig:real-results} reports the success rates under in-distribution,
out-of-distribution (unseen object colors and initial poses) and
dark-background conditions. ACiT is the strongest or tied-strongest method in
every condition and task, and its margin grows under the two shifts: with a dark
background it still picks cube at 83\%, against less than 40\% for the baselines.

\subsection{Transfer to a Large VLA}
\label{sec:vla}

\begin{figure}[t]
\centering
\includegraphics[width=\columnwidth]{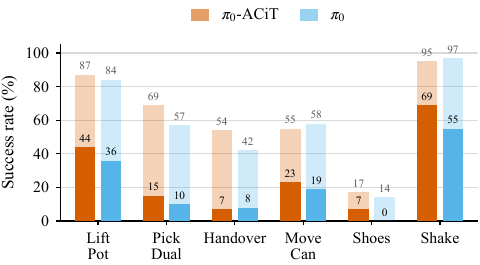}
\caption{Transferring the conditioning design to a large VLA on RoboTwin~2.0.
Success rate of $\pi_0$-ACiT (left bar of each task) and $\pi_0$ (right bar).
The pale bar is the easy rate and the solid bar drawn on top of it is the
randomized-hard rate, so the pale cap is the drop caused by randomization.
Numbers above the pale bars are easy rates, numbers on the solid bars hard rates.}
\label{fig:vla-sim}
\end{figure}

\begin{figure}[t]
\centering
\includegraphics[width=\columnwidth]{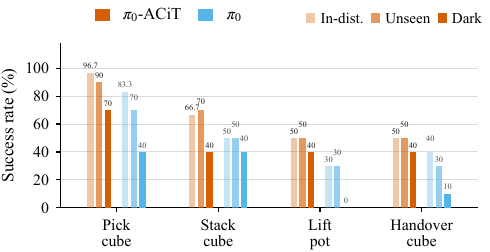}
\caption{Real-world $\pi_0$/$\pi_0$-ACiT transfer on the B601 platform. Each
task has six bars: $\pi_0$-ACiT under in-distribution, unseen-object/background,
and dark-background conditions (light to dark orange), followed by $\pi_0$ under
the same conditions (light to dark blue).}
\label{fig:real-vla}
\end{figure}

We apply ACiT's conditioning to the post-training of the mainstream
$\pi_0$ model~\citep{pi0}, which yields $\pi_0$-ACiT. Both models are fine-tuned
end-to-end per task, with the PaliGemma VLM~\citep{paligemma} and the flow-matching expert keeping
their architecture and objective. Only $\pi_0$'s single-frame proprioceptive
token is replaced by $\mathbf{G}_t$ from the Inverted State Encoding and
State-Visual Grounding of ACiT.

\subsubsection{Simulation Benchmarks}
Fig.~\ref{fig:vla-sim} compares the two models on six RoboTwin tasks. The gains land
on the same group of tasks as the ACT-family results of Fig.~\ref{fig:main-results}:
both backbones improve where one frame cannot determine each joint's action on its
own (Pick Dual Bottles, Handover Block, Place Dual Shoes) and stay flat where it can
(Lift Pot, Move Can Pot, Shake Bottle). Two backbones that differ in size, visual
prior and action head agree on which tasks benefit, so the gain comes from the
conditioning interface rather than from a property of a backbone. On the randomized
hard split it moves to Shake Bottle and Lift Pot, whose actions depend on velocity and
on the phase relation between joints, a relation that exists only in the history. Randomization removes the same fraction for both models, so the gain is not a
tolerance to visual shift.

\subsubsection{Real-World Transfer}
Fig.~\ref{fig:real-vla} repeats the comparison on the dual-arm B601 platform.
$\pi_0$-ACiT improves over $\pi_0$ on average across the four tasks and three
conditions (57.0 against 39.4). The gain is
largest under the dark-background shift, where $\pi_0$ nearly fails on the two
dual-arm tasks and $\pi_0$-ACiT still completes them. In distribution and under the unseen
shift $\pi_0$-ACiT matches the ACT-based ACiT, so the interface transfers to a large pretrained
backbone at a lower cost per action (Table~\ref{tab:efficiency}). Averaged over the 12
RoboTwin settings (6 tasks, 2 splits) and the 12 real-world settings (4 tasks, 3
conditions), $\pi_0$-ACiT raises the success rate over $\pi_0$ by 26.9 points.

\subsection{Ablation}
\label{sec:ablation}

Table~\ref{tab:ablation} removes one module at a time, keeping the history
length, the encoder, the protocol and the action head fixed. \emph{Equal fusion}
self-attends visual and joint tokens together. \emph{w/o joint attn} keeps the
per-joint 1D convolution but drops the attention among joint tokens, and
\emph{flat state} drops both. \emph{Flat decode} replaces per-joint decoding with
one shared query.

Removing any one module hurts on every benchmark, so the three are complementary
rather than redundant. The conditioning relation costs most, the state-side
variants next and the action-side factorization least, in five of the six settings.
The exception is the unseen condition, where \emph{flat state} falls further than
\emph{equal fusion}, so under unreliable vision the per-joint state representation
matters more than the fusion relation.

\begin{table}[h!]
\caption{Module ablation (mean success rate \%). Small numbers give the change
relative to the full model, per-task rates in Appendix~\xref{app:ablation-full}.}
\label{tab:ablation}
\centering
\scriptsize
\renewcommand{\arraystretch}{1.05}
\setlength{\tabcolsep}{3pt}
\resizebox{\columnwidth}{!}{%
\begin{tabular}{l c|cc|ccc}
\toprule
Variant & LIBERO & \multicolumn{2}{c|}{RoboTwin} & \multicolumn{3}{c}{Real world} \\
\cmidrule(lr){2-2}\cmidrule(lr){3-4}\cmidrule(lr){5-7}
 & & easy & hard & ID & unseen & dark \\
\midrule
\shortstack[l]{ACiT (full)\\\phantom{x}} & \shortstack{\makebox[2.4em][r]{79.0}\\\makebox[2.4em][r]{\phantom{x}}} & \shortstack{\makebox[2.4em][r]{62.7}\\\makebox[2.4em][r]{\phantom{x}}} & \shortstack{\makebox[2.4em][r]{14.0}\\\makebox[2.4em][r]{\phantom{x}}} & \shortstack{\makebox[2.4em][r]{60.0}\\\makebox[2.4em][r]{\phantom{x}}} & \shortstack{\makebox[2.4em][r]{34.1}\\\makebox[2.4em][r]{\phantom{x}}} & \shortstack{\makebox[2.4em][r]{44.1}\\\makebox[2.4em][r]{\phantom{x}}} \\
\shortstack[l]{equal fusion\\\phantom{x}} & \shortstack{\makebox[2.4em][r]{58.3}\\\makebox[2.4em][r]{\tiny\textcolor{gray}{$-20.7$}}} & \shortstack{\makebox[2.4em][r]{52.3}\\\makebox[2.4em][r]{\tiny\textcolor{gray}{$-10.4$}}} & \shortstack{\makebox[2.4em][r]{5.0}\\\makebox[2.4em][r]{\tiny\textcolor{gray}{$-9.0$}}} & \shortstack{\makebox[2.4em][r]{48.4}\\\makebox[2.4em][r]{\tiny\textcolor{gray}{$-11.6$}}} & \shortstack{\makebox[2.4em][r]{17.5}\\\makebox[2.4em][r]{\tiny\textcolor{gray}{$-16.6$}}} & \shortstack{\makebox[2.4em][r]{13.3}\\\makebox[2.4em][r]{\tiny\textcolor{gray}{$-30.8$}}} \\
\shortstack[l]{w/o joint attn\\\phantom{x}} & \shortstack{\makebox[2.4em][r]{69.3}\\\makebox[2.4em][r]{\tiny\textcolor{gray}{$-9.7$}}} & \shortstack{\makebox[2.4em][r]{57.0}\\\makebox[2.4em][r]{\tiny\textcolor{gray}{$-5.7$}}} & \shortstack{\makebox[2.4em][r]{9.8}\\\makebox[2.4em][r]{\tiny\textcolor{gray}{$-4.2$}}} & \shortstack{\makebox[2.4em][r]{52.5}\\\makebox[2.4em][r]{\tiny\textcolor{gray}{$-7.5$}}} & \shortstack{\makebox[2.4em][r]{23.3}\\\makebox[2.4em][r]{\tiny\textcolor{gray}{$-10.8$}}} & \shortstack{\makebox[2.4em][r]{28.3}\\\makebox[2.4em][r]{\tiny\textcolor{gray}{$-15.8$}}} \\
\shortstack[l]{flat state\\\phantom{x}} & \shortstack{\makebox[2.4em][r]{65.3}\\\makebox[2.4em][r]{\tiny\textcolor{gray}{$-13.7$}}} & \shortstack{\makebox[2.4em][r]{53.3}\\\makebox[2.4em][r]{\tiny\textcolor{gray}{$-9.4$}}} & \shortstack{\makebox[2.4em][r]{7.0}\\\makebox[2.4em][r]{\tiny\textcolor{gray}{$-7.0$}}} & \shortstack{\makebox[2.4em][r]{49.2}\\\makebox[2.4em][r]{\tiny\textcolor{gray}{$-10.8$}}} & \shortstack{\makebox[2.4em][r]{15.0}\\\makebox[2.4em][r]{\tiny\textcolor{gray}{$-19.1$}}} & \shortstack{\makebox[2.4em][r]{21.6}\\\makebox[2.4em][r]{\tiny\textcolor{gray}{$-22.5$}}} \\
\shortstack[l]{flat decode\\\phantom{x}} & \shortstack{\makebox[2.4em][r]{75.0}\\\makebox[2.4em][r]{\tiny\textcolor{gray}{$-4.0$}}} & \shortstack{\makebox[2.4em][r]{60.0}\\\makebox[2.4em][r]{\tiny\textcolor{gray}{$-2.7$}}} & \shortstack{\makebox[2.4em][r]{11.8}\\\makebox[2.4em][r]{\tiny\textcolor{gray}{$-2.2$}}} & \shortstack{\makebox[2.4em][r]{54.2}\\\makebox[2.4em][r]{\tiny\textcolor{gray}{$-5.8$}}} & \shortstack{\makebox[2.4em][r]{25.8}\\\makebox[2.4em][r]{\tiny\textcolor{gray}{$-8.3$}}} & \shortstack{\makebox[2.4em][r]{41.6}\\\makebox[2.4em][r]{\tiny\textcolor{gray}{$-2.5$}}} \\
\bottomrule
\end{tabular}}
\end{table}

\subsection{Robustness}
\label{sec:robust}

\runinhead{Training Seeds.}%
Two further seeds leave the task ranking unchanged and move each task by at most
eight points. Averaged over the tasks the success rate is 62.7 $\pm$ 0.6, so the
aggregate in Fig.~\ref{fig:main-results} does not depend on the seed, while a single
task carries the few points of spread that 100 episodes leave.

\begin{table}[t]
\centering
\caption{Training-seed variance of ACiT (easy split, success rate over 100 episodes,
hence integer).}
\label{tab:seeds}
\small
\setlength{\tabcolsep}{3pt}
\begin{tabular}{@{}lcccc@{}}
\toprule
Task & seed 1000 & seed 1001 & seed 1002 & mean\,$\pm$\,std \\
\midrule
Lift Pot        & 55 & 56 & 54 & $55.0\pm1.0$ \\
Pick Dual       & 61 & 63 & 59 & $61.0\pm2.0$ \\
Handover        & 69 & 69 & 69 & $69.0\pm0.0$ \\
Move Can        & 64 & 60 & 68 & $64.0\pm4.0$ \\
Place Dual Shoes & 41 & 35 & 38 & $38.0\pm3.0$ \\
Shake Bottle    & 88 & 89 & 90 & $89.0\pm1.0$ \\
\bottomrule
\end{tabular}
\end{table}

\begin{figure}[t]
\centering
\includegraphics[width=\columnwidth]{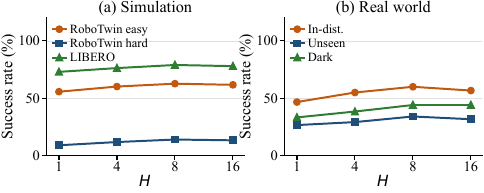}
\caption{History length of the conditioning input. (a) Simulation: RoboTwin easy and
hard, and LIBERO. (b) Real world: in-distribution, unseen and dark-background. All
curves peak at $H{=}8$ or tie it at 16; the per-task rates are in
Appendix~\xref{app:ablation}.}
\label{fig:hist}
\end{figure}

\runinhead{History Length.}%
Figure~\ref{fig:hist} shows a flat optimum: performance peaks at eight frames, changes
little between four and sixteen, and is clearly worse with a single frame, whose cost
is largest under the dark-background shift. We keep eight frames throughout.

\subsection{Efficiency}
\label{sec:efficiency}

Table~\ref{tab:efficiency} reports inference cost and model size. ACiT adds only
8.6\,M parameters over ACT (92.6\,M against 84.0\,M) and keeps its one-step sampler,
at 0.48\,ms per action, while DP and FM pay ten integrator steps at a
3.2$\times$ larger model. The same holds after transfer: $\pi_0$-ACiT needs
4.73\,ms per action against 5.93\,ms for $\pi_0$, at the same ten steps and
essentially the same size (4.05\,B against 4.03\,B), so the reduction comes from
the interface, not from a cheaper sampler.

\begin{table}[t]
\centering
\caption{Inference cost on one RTX~3090, at the $240\times320$ observation
resolution used throughout the RoboTwin experiments. \emph{Steps} counts integrator
steps (network evaluations in brackets), and the last two columns are per executed
action.}
\label{tab:efficiency}
\small
\setlength{\tabcolsep}{2pt}
\begin{tabular*}{\columnwidth}{@{\extracolsep{\fill}}lccrr@{}}
\toprule
Method & Params & Steps & ms/act. & Control (Hz) \\
\midrule
ACT & 84.0\,M & 1 (1) & 0.36 & 2794 \\
DP & 293.2\,M & 10 (10) & 2.12 & 471 \\
FM & 293.2\,M & 10 (20) & 3.74 & 267 \\
ACiT & 92.6\,M & 1 (1) & 0.48 & 2064 \\
\midrule
$\pi_0$ & 4.03\,B & 10 (10) & 5.93 & 169 \\
$\pi_0$-ACiT & 4.05\,B & 10 (10) & 4.73 & 211 \\
\bottomrule
\end{tabular*}
\end{table}

\section{Conclusion}
\label{sec:conclusion}

We presented ACiT, which treats historical joint coupling as the primary condition
and vision as a later calibration: each joint carries its own history,
retrieves its own visual evidence and decodes its own trajectory. It improves most
LIBERO, RoboTwin~2.0 and real-world tasks, and transfers to a pretrained VLA at
lower cost.

\runinhead{Limitations.}%
ACiT is instantiated only with a continuous action head, so it does not cover the
discrete action tokens that autoregressive VLAs rely on. Extending it there needs
demonstration data beyond what our dual-arm platform provides, which we leave to
future work.

\par\medskip

\section*{Acknowledgments}

We thank Shenzhen Zhibo Data Technology Co., Ltd. for the professional
collection of the real-robot data used in this work.

\bibliography{references}
\bibliographystyle{plainnat}

\clearpage
\clearpage

\ifdefined\WITHAPPENDIX
\setcounter{dbltopnumber}{5}
\renewcommand{\dbltopfraction}{1.0}
\setcounter{topnumber}{4}
\raggedbottom
\appendix
\suppressfloats[t]
\makeatletter
\setlength{\@fptop}{0pt}
\setlength{\@fpsep}{8pt plus 2pt}
\setlength{\@fpbot}{0pt plus 1fil}
\setlength{\@dblfptop}{0pt}\setlength{\@dblfpsep}{8pt plus 2pt}\setlength{\@dblfpbot}{0pt plus 1fil}
\makeatother
\setlength{\textfloatsep}{10pt plus 2pt minus 2pt}
\setlength{\dbltextfloatsep}{10pt plus 2pt minus 2pt}
\setlength{\dblfloatsep}{8pt plus 1pt minus 1pt}
\setlength{\floatsep}{8pt plus 1pt minus 1pt}
\setlength{\intextsep}{8pt plus 1pt minus 1pt}

\section{Experimental Setup}
\label{app:setup}

\subsection{Simulation Benchmarks}
\label{app:sim-tasks}

RoboTwin~2.0 is a bimanual platform with two 6-DoF arms and parallel grippers,
simulated for table-top manipulation. We use six tasks, shown in
Fig.~\ref{fig:sim-tasks}. \emph{Lift Pot} lifts a pot by both handles and holds
it; \emph{Pick Dual Bottles} grasps and lifts two bottles at once;
\emph{Handover Block} passes a block from one arm to the other;
\emph{Move Can Pot} lays the can against the pot; \emph{Place Dual Shoes} places two
shoes into a box; \emph{Shake Bottle} lifts a bottle and shakes it. Each task
comes with an easy split (\texttt{demo\_clean}) and a randomized split
(\texttt{demo\_randomized}) in which object poses, object colors and the scene
background are randomized. All numbers are 100 episodes per task, and the
success criterion is the one shipped with the benchmark, listed in
Table~\ref{tab:criteria}.

\begin{table}[t]
\centering
\caption{Success conditions of the six RoboTwin tasks, as implemented by the
benchmark. A rollout counts as a success only when every condition of its task
holds at the same step.}
\label{tab:criteria}
\small
\setlength{\tabcolsep}{4pt}
\begin{tabular}{@{}p{0.30\columnwidth}p{0.62\columnwidth}@{}}
\toprule
Task & Success condition \\
\midrule
Lift Pot & pot above 0.82\,m and upright, both TCPs within 3\,cm of its handles \\
Pick Dual Bottles & both bottles within 10\,cm of their targets and above 0.89\,m \\
Handover Block & block within 3\,cm of the target pose and right gripper open \\
Move Can Pot & can tipped on its side beside the pot ($<$20\,cm, 3.5\,cm, $15^\circ$), at table height, both grippers open \\
Place Dual Shoes & both shoes at the target pose (5\,cm, 0.07 quaternion) and both grippers open \\
Shake Bottle & bottle raised above 0.8\,m \\
\bottomrule
\end{tabular}
\end{table}

LIBERO evaluates single-arm table-top manipulation with language instructions. We
use three suites: LIBERO-10 (ten long-horizon tasks), LIBERO-Object (objects
unseen in training) and LIBERO-Spatial (spatial relations), with 100 episodes per
suite.

\subsection{Implementation Details}
\label{app:impl}

The ACT-family models share one protocol: 50{,}000 training steps, batch size
16, action chunk 50 with 50 executed steps, seed 1000, and AdamW~\citep{adamw}. The
visual backbone is a ResNet-18~\citep{resnet}. All models are implemented in
PyTorch~\citep{pytorch} with LeRobot~\citep{lerobot}, and the simulator is
MuJoCo~\citep{mujoco}. For run-to-run variation we follow the protocol of
ACT~\citep{act}, which reports three training seeds on simulation and a single
trained policy on hardware, where every additional seed needs a full physical
re-evaluation. Table~\ref{tab:seeds} of the main paper reports the same sweep. ACiT uses a state history of $H{=}8$ frames. ACT
conditions on the current state only; DP and FM take $H{=}2$
frames and flatten them into the global condition, which is the setting their
original papers use. DP and FM execute 32 steps in the
\texttt{50k\_final} runs and 10 in the \texttt{5task} runs.

Architecturally, ACiT keeps the ACT transformer ($d{=}512$, eight heads, FFN width
3200, four encoder and seven decoder layers, 32-d CVAE latent) and adds a
two-layer state encoder followed by a two-layer cross-attention block. The
$\pi_0$ backbone is left intact; $\pi_0$-ACiT adds a two-layer state encoder and a
single grounding layer to it and keeps the flow-matching action expert unchanged.

For the VLA transfer (Sec.~\ref{sec:vla}), we initialize $\pi_0$-ACiT from
\texttt{pi0\_base}. The PaliGemma~\citep{paligemma} backbone and the flow-matching action expert
keep their original architecture, and $\pi_0$'s single-frame proprioceptive token
is replaced by the two conditioning modules of ACiT. The \emph{Inverted State
Encoding} encodes each joint's eight-frame history with a 1D convolution and a
linear projection into a joint token, and forms the proprioceptive prior by
inter-joint self-attention. \emph{State-Visual Grounding} then applies a
joint-level cross-attention over the visual tokens, producing $\mathbf{G}_t$ as
the proprioceptive condition of the action expert. The \emph{Inverted Action
Chunking} is not transferred, and the flow-matching objective is unchanged, so
the action head keeps its architecture and capacity and only the conditioning
side changes. Both models are trained per task with full fine-tuning, batch size
32, for 50k steps on two A100 GPUs.

Evaluation follows the benchmark defaults: $240\times320$ on RoboTwin for every
model, with 100
episodes per task, $256\times256$ on LIBERO with 100 episodes per suite, and 30
rollouts per task and condition on the real-world platform, where training uses
50 demonstrations per task, the same recipe as above, and all rollouts run on a
single RTX~3090.

\begin{figure*}[t!]
\centering
\includegraphics[width=\textwidth]{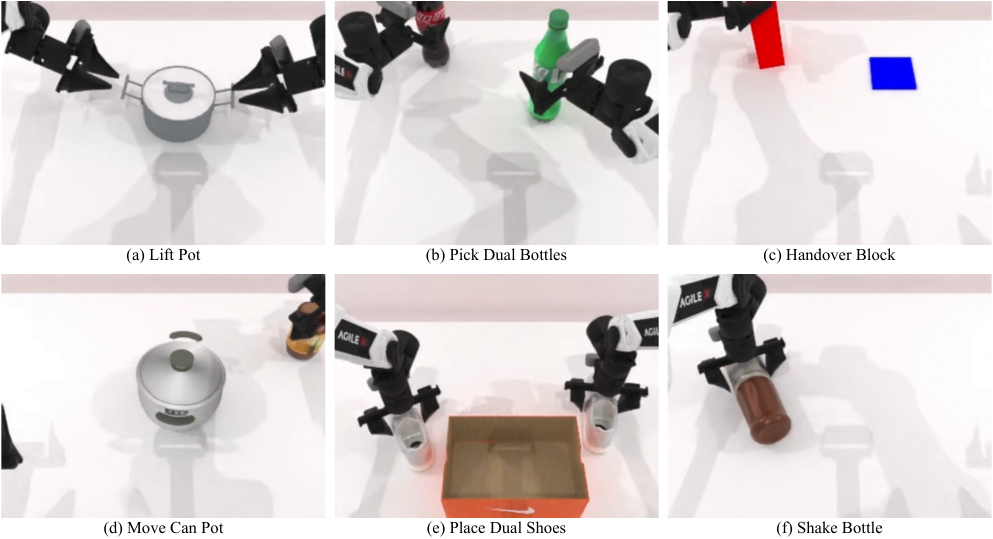}
\caption{The six RoboTwin~2.0 tasks used in Sec.~\ref{sec:main}, viewed from the
head camera. Both arms are visible in each panel; the randomized split varies
object poses, colors and the background of the same scenes.}
\label{fig:sim-tasks}
\end{figure*}

\section{Ablations, Sensitivity and Cost}
\label{app:ablation}

\subsection{Module Ablation}
\label{app:ablation-full}

Tables~\ref{tab:ablation-sim} and~\ref{tab:ablation-real} give the per-task rates
behind Table~\ref{tab:ablation}. The three removals do not fail in the same
places. \emph{Equal fusion} costs the most on Handover Block and Place Dual
Shoes, the two tasks whose success turns on coupling between joints, and it is
almost free on Move Can Pot. \emph{Flat state} costs a nearly constant amount on
every task, which is what one expects when the model loses the per-joint state
representation rather than a specific coupling. \emph{Flat decode} is the
opposite: it loses on Pick Dual Bottles and Place Dual Shoes, where each joint
needs its own trajectory, and it gains on Handover Block and Move Can Pot, where
one shared trajectory keeps the joints in phase.

\begin{table*}[t]
\caption{Per-task module ablation on the simulation benchmarks (success rate,
\%, 100 episodes per task).}
\label{tab:ablation-sim}
\centering
\scriptsize
\setlength{\tabcolsep}{3.2pt}
\begin{tabular*}{\textwidth}{@{\extracolsep{\fill}}lcccccc c cccccc c ccc c@{}}
\toprule
& \multicolumn{7}{c}{\textbf{RoboTwin (easy)}} & \multicolumn{7}{c}{\textbf{RoboTwin (hard)}} & \multicolumn{4}{c}{\textbf{LIBERO}} \\
\cmidrule(lr){2-8}\cmidrule(lr){9-15}\cmidrule(lr){16-19}
Variant & Lift & Pick & Hand. & Move & Shoes & Shake & Mean & Lift & Pick & Hand. & Move & Shoes & Shake & Mean & 10 & Obj. & Spa. & Mean \\
\midrule
ACiT & 55 & 61 & 69 & 64 & 38 & 89 & 62.7 & 4 & 14 & 0 & 19 & 0 & 47 & 14.0 & 50 & 99 & 88 & 79.0 \\
equal fusion & 56 & 50 & 30 & 68 & 22 & 88 & 52.3 & 1 & 2 & 0 & 2 & 0 & 25 & 5.0 & 30 & 73 & 72 & 58.3 \\
w/o joint attn & 39 & 60 & 69 & 62 & 30 & 82 & 57.0 & 3 & 6 & 0 & 12 & 0 & 38 & 9.8 & 38 & 90 & 80 & 69.3 \\
flat state & 48 & 50 & 58 & 56 & 28 & 80 & 53.3 & 2 & 3 & 0 & 7 & 0 & 30 & 7.0 & 35 & 85 & 76 & 65.3 \\
flat decode & 54 & 45 & 79 & 74 & 23 & 85 & 60.0 & 5 & 8 & 1 & 16 & 0 & 41 & 11.8 & 45 & 95 & 85 & 75.0 \\
\bottomrule
\end{tabular*}
\end{table*}

\begin{table*}[t]
\caption{Per-task module ablation on the real-world platform (success rate, \%;
30 rollouts per cell).}
\label{tab:ablation-real}
\centering
\scriptsize
\setlength{\tabcolsep}{3.2pt}
\begin{tabular*}{\textwidth}{@{\extracolsep{\fill}}l cccc c cccc c cccc c@{}}
\toprule
& \multicolumn{5}{c}{\textbf{In-distribution}} & \multicolumn{5}{c}{\textbf{Unseen}} & \multicolumn{5}{c}{\textbf{Dark background}} \\
\cmidrule(lr){2-6}\cmidrule(lr){7-11}\cmidrule(lr){12-16}
Variant & Pick & Stack & Pot & Hand. & Mean & Pick & Stack & Pot & Hand. & Mean & Pick & Stack & Pot & Hand. & Mean \\
\midrule
ACiT & 93.3 & 63.3 & 33.3 & 50.0 & 60.0 & 53.3 & 23.3 & 16.7 & 43.3 & 34.1 & 83.3 & 43.3 & 16.7 & 33.3 & 44.1 \\
equal fusion & 76.7 & 56.7 & 23.3 & 36.7 & 48.4 & 40.0 & 6.7 & 6.7 & 16.7 & 17.5 & 30.0 & 13.3 & 0.0 & 10.0 & 13.3 \\
w/o joint attn & 80.0 & 56.7 & 30.0 & 43.3 & 52.5 & 43.3 & 13.3 & 13.3 & 23.3 & 23.3 & 53.3 & 23.3 & 13.3 & 23.3 & 28.3 \\
flat state & 70.0 & 56.7 & 23.3 & 46.7 & 49.2 & 33.3 & 6.7 & 6.7 & 13.3 & 15.0 & 43.3 & 23.3 & 6.7 & 13.3 & 21.6 \\
flat decode & 83.3 & 46.7 & 30.0 & 56.7 & 54.2 & 43.3 & 13.3 & 13.3 & 33.3 & 25.8 & 73.3 & 30.0 & 20.0 & 43.3 & 41.6 \\
\bottomrule
\end{tabular*}
\end{table*}

\subsection{Sensitivity to History Length}
\label{app:history}

History length is a hyperparameter of the conditioning interface rather than a
module, so we report it separately. Tables~\ref{tab:hist-sim}
and~\ref{tab:hist-real} vary the number of proprioceptive frames and keep
everything else fixed. The same pattern appears everywhere: one frame is the
weakest, four recovers most of the gap, eight is best on RoboTwin and on the real
platform and matches four on LIBERO, and sixteen is slightly worse, which is what
one expects when extra frames add redundancy but no new dynamics.

\begin{table*}[t]
\caption{History length on the simulation benchmarks (success rate, \%, 100
episodes per task). $H$ is the history length, in frames.}
\label{tab:hist-sim}
\centering
\scriptsize
\setlength{\tabcolsep}{3pt}
\begin{tabular*}{\textwidth}{@{\extracolsep{\fill}}l cccccc c cccccc c ccc c@{}}
\toprule
& \multicolumn{7}{c}{\textbf{RoboTwin (easy)}} & \multicolumn{7}{c}{\textbf{RoboTwin (hard)}} & \multicolumn{4}{c}{\textbf{LIBERO}} \\
\cmidrule(lr){2-8}\cmidrule(lr){9-15}\cmidrule(lr){16-19}
$H$ & Lift & Pick & Hand. & Move & Shoes & Shake & Mean & Lift & Pick & Hand. & Move & Shoes & Shake & Mean & 10 & Obj. & Spa. & Mean \\
\midrule
1 & 52 & 45 & 60 & 58 & 34 & 85 & 55.7 & 2 & 5 & 0 & 12 & 0 & 35 & 9.0 & 46 & 91 & 82 & 73.0 \\
4 & 54 & 56 & 66 & 62 & 36 & 87 & 60.2 & 3 & 10 & 0 & 16 & 0 & 42 & 11.8 & 51 & 99 & 88 & 79.3 \\
8 & 55 & 61 & 69 & 64 & 38 & 89 & 62.7 & 4 & 14 & 0 & 19 & 0 & 47 & 14.0 & 50 & 99 & 88 & 79.0 \\
16 & 54 & 60 & 68 & 63 & 37 & 88 & 61.7 & 4 & 13 & 0 & 18 & 0 & 45 & 13.3 & 49 & 98 & 87 & 78.0 \\
\bottomrule
\end{tabular*}
\end{table*}

\begin{table*}[t]
\caption{History length on the real-world platform (success rate, \%; 30
rollouts per cell).}
\label{tab:hist-real}
\centering
\scriptsize
\setlength{\tabcolsep}{4pt}
\begin{tabular*}{\textwidth}{@{\extracolsep{\fill}}l cccc c cccc c cccc c@{}}
\toprule
& \multicolumn{5}{c}{\textbf{In-distribution}} & \multicolumn{5}{c}{\textbf{Unseen}} & \multicolumn{5}{c}{\textbf{Dark background}} \\
\cmidrule(lr){2-6}\cmidrule(lr){7-11}\cmidrule(lr){12-16}
$H$ & Pick & Stack & Pot & Hand. & Mean & Pick & Stack & Pot & Hand. & Mean & Pick & Stack & Pot & Hand. & Mean \\
\midrule
1 & 76.7 & 46.7 & 23.3 & 40.0 & 46.7 & 43.3 & 16.7 & 13.3 & 33.3 & 26.6 & 70.0 & 33.3 & 10.0 & 20.0 & 33.3 \\
4 & 86.7 & 56.7 & 30.0 & 46.7 & 55.0 & 46.7 & 20.0 & 13.3 & 36.7 & 29.2 & 76.7 & 36.7 & 13.3 & 26.7 & 38.4 \\
8 & 93.3 & 63.3 & 33.3 & 50.0 & 60.0 & 53.3 & 23.3 & 16.7 & 43.3 & 34.1 & 83.3 & 43.3 & 16.7 & 33.3 & 44.1 \\
16 & 90.0 & 60.0 & 30.0 & 46.7 & 56.7 & 50.0 & 20.0 & 16.7 & 40.0 & 31.7 & 83.3 & 43.3 & 16.7 & 33.3 & 44.1 \\
\bottomrule
\end{tabular*}
\end{table*}

\subsection{Inference Cost}
\label{app:efficiency}

Cost is measured on a single RTX~3090 (24\,GB) with batch size 1 and fp16, and
each entry of Table~\ref{tab:efficiency} averages 100 calls after 20 warm-up
calls on the RoboTwin \emph{Lift Pot} checkpoints at 50k steps. The table reports the
cost per executed action, which divides one inference by the number of actions run
without re-planning, i.e., what a closed-loop controller pays. The figures below are
per inference, so that samplers can be compared directly. Step counts alone do not
fix the cost, because
integrators differ in how many network evaluations (NFE) they spend per step. DP and FM share the same U-Net and the same parameter count
but not the same sampler: DP takes one evaluation per DDIM step, whereas FM
integrates with a midpoint rule that evaluates the network twice per step, so its
ten steps cost twenty evaluations (119.8\,ms per inference) and match DP at twenty
steps (120.5\,ms); at an equal budget of ten evaluations the two coincide
(67.5 against 68.1\,ms). We therefore report both steps and NFE, and read cost
differences across families as sampler differences unless the NFE budget is held
fixed.

\subsection{Seed Variance}
\label{app:seed}

Table~\ref{tab:seeds} of the main paper repeats the training of ACiT and ACT on
RoboTwin with two further seeds. The evaluation episodes are shared across runs, so
the only source of variation is the training itself, and the spread across seeds stays
small relative to the gap between the two models.

\section{Analysis}
\label{app:analysis}

\subsection{Failure Cases}
\label{app:failure}

\begin{figure*}[t]
\centering
\includegraphics[width=\textwidth]{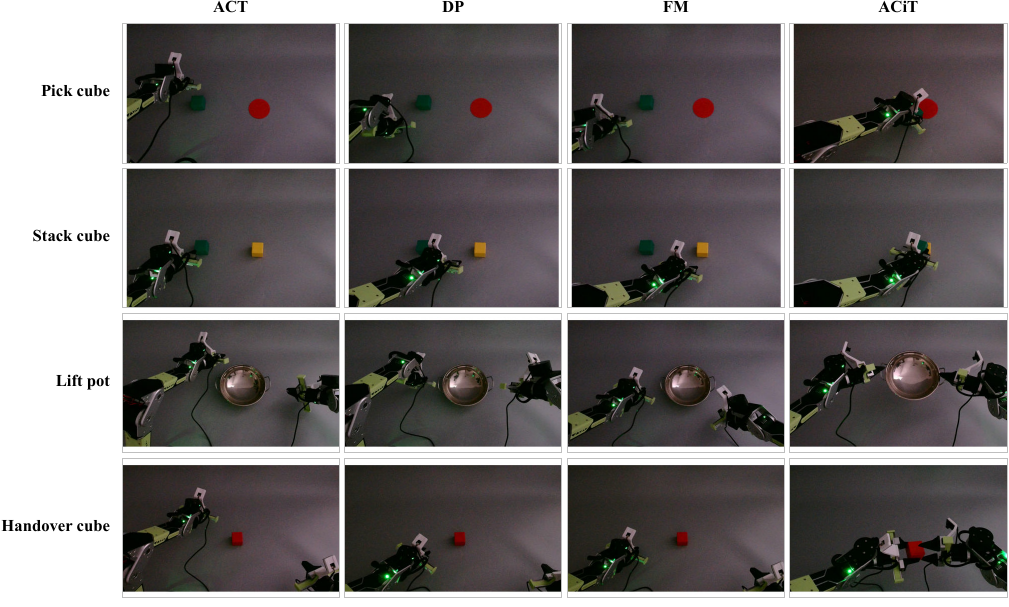}
\caption{Out-of-distribution dark background. ACT, DP, FM and ACiT on the four tasks under the background shift.}
\label{fig:ood-background}
\end{figure*}

\begin{figure*}[t]
\centering
\includegraphics[width=\textwidth]{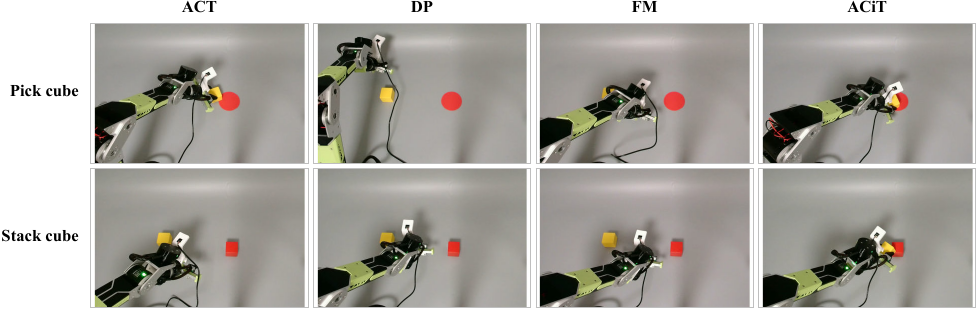}
\caption{Out-of-distribution unseen initial state and color, on Pick cube and Stack cube.}
\label{fig:ood-color}
\end{figure*}

Figs.~\ref{fig:ood-background} and~\ref{fig:ood-color} also show how the methods
fail once the scene is shifted. The baselines fail at the grasp itself: under a
changed background, ACT and FM approach the cube with an offset and close on
empty space, and DP pushes it away before closing; under a changed object color
the same offset appears, which is the behavior expected when a single frame has
to carry both the pose and the appearance. ACiT keeps the grasp in the same
frames and fails later, when placing the object. Its remaining failures are
therefore concentrated in placement precision and in the phase of a handover,
not in grasp stability, which is consistent with Fig.~\ref{fig:real-results}: the gap to the baselines is
smallest on the two tasks whose success is dominated by the grasp and largest
under the dark shift.

Figs.~\ref{fig:qual-shoes}--\ref{fig:qual-libero} replay one episode of each task
under all four policies, with the checker outcome and the number of steps used
marked on the last frame; the success conditions are listed in
Table~\ref{tab:criteria}; every row of a figure shares the same initial state. On
Place Dual Shoes success requires both shoes inside the box with the target
orientation and both grippers open, which only ACiT satisfies. On Pick Dual
Bottles ACiT lifts the bottle within a fraction of the step budget while the
baselines never leave the table. On Lift Pot ACiT and DP both succeed, with ACiT
finishing in fewer steps, while ACT and FM fail. On Move Can Pot only ACiT
brings the can to the pot. On Shake Bottle DP never lifts the bottle, and on
LIBERO-Spatial all four policies succeed with ACiT taking the fewest steps.

\begin{figure*}[t]
\centering
\includegraphics[width=\textwidth]{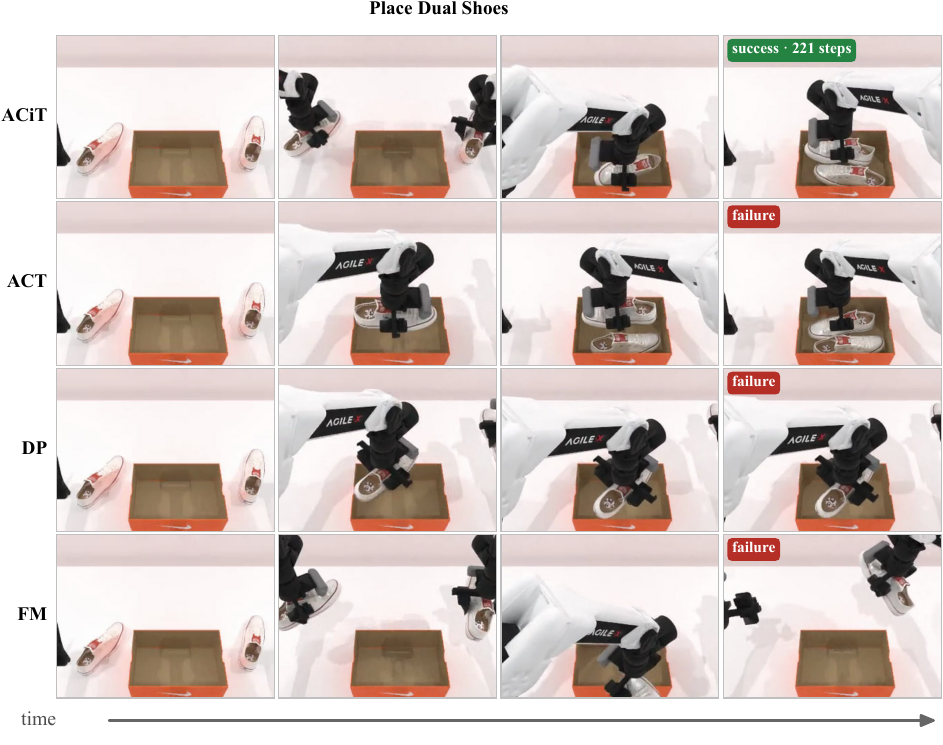}
\caption{Qualitative comparison on Place Dual Shoes, same task and episode under four
policies. The outcome and the number of steps are marked on the last frame. Only ACiT
leaves both shoes in the box and releases them; ACT keeps a shoe in its closed gripper,
and DP and FM stop with one shoe outside the box.}
\label{fig:qual-shoes}
\end{figure*}

\begin{figure*}[t]
\centering
\includegraphics[width=\textwidth]{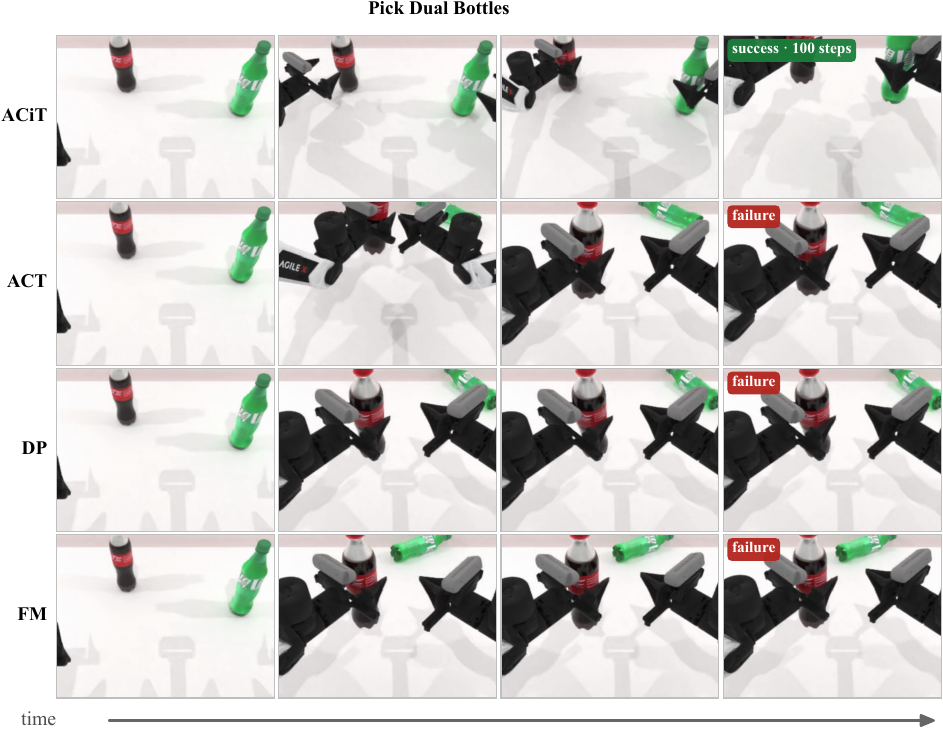}
\caption{Qualitative comparison on Pick Dual Bottles. ACiT lifts the bottle clear of the
table and finishes well inside the step budget, while ACT, DP and FM leave both bottles
in place.}
\label{fig:qual-pick}
\end{figure*}

\begin{figure*}[t]
\centering
\includegraphics[width=\textwidth]{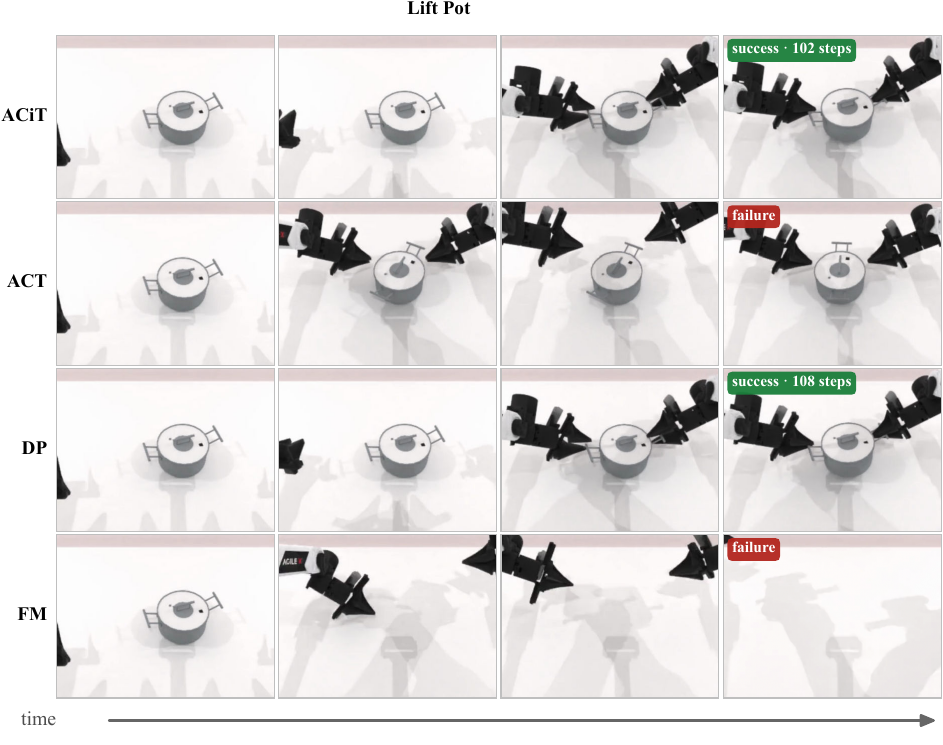}
\caption{Qualitative comparison on Lift Pot. ACiT and DP both raise the pot, with ACiT
finishing in fewer steps, while ACT and FM fail to lift it.}
\label{fig:qual-lift}
\end{figure*}

\begin{figure*}[t]
\centering
\includegraphics[width=\textwidth]{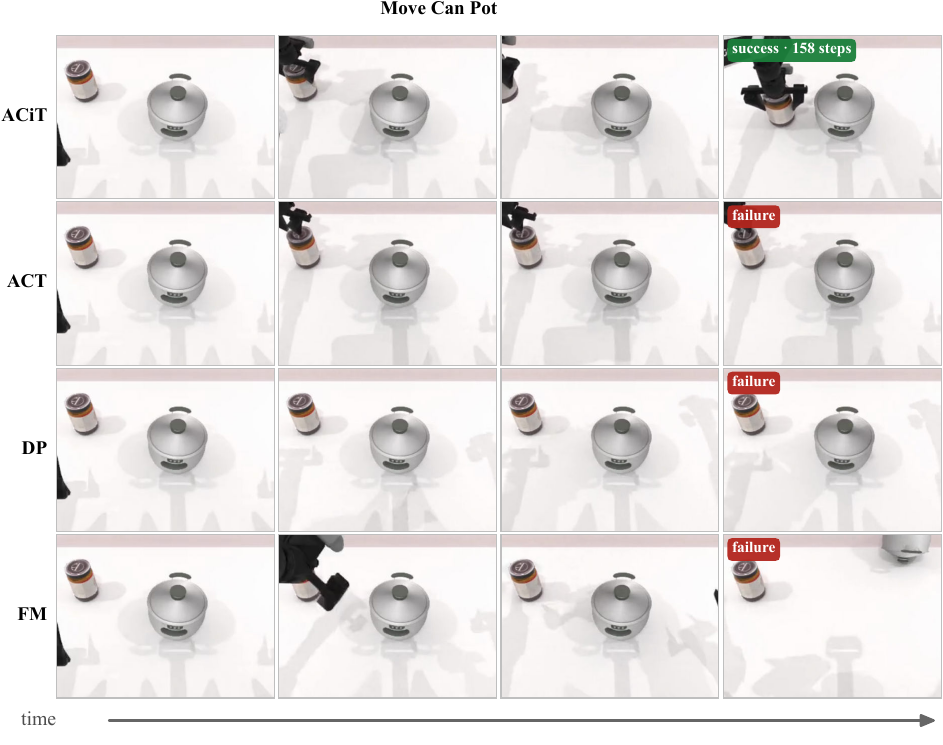}
\caption{Qualitative comparison on Move Can Pot. ACiT places the can against the pot and
finishes in 158 steps; ACT never moves it into place, DP does not approach it, and FM
knocks it away from its start position.}
\label{fig:qual-can}
\end{figure*}

\begin{figure*}[t]
\centering
\includegraphics[width=\textwidth]{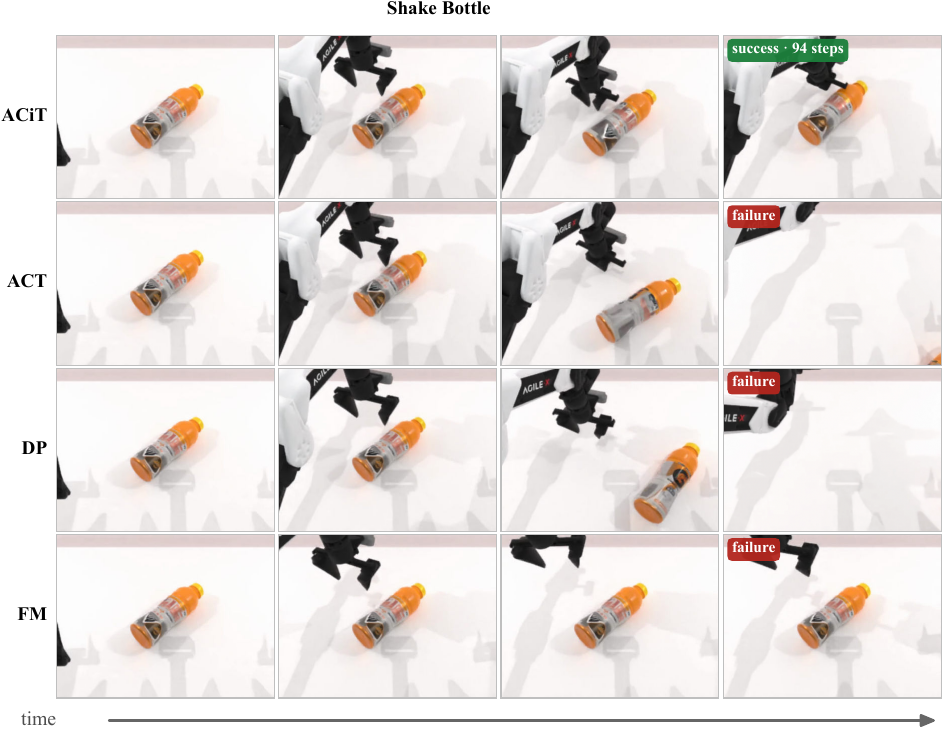}
\caption{Qualitative comparison on Shake Bottle. ACiT, ACT and FM shake the bottle, while
DP never lifts it and is marked failure.}
\label{fig:qual-shake}
\end{figure*}

\begin{figure*}[t]
\centering
\includegraphics[width=\textwidth]{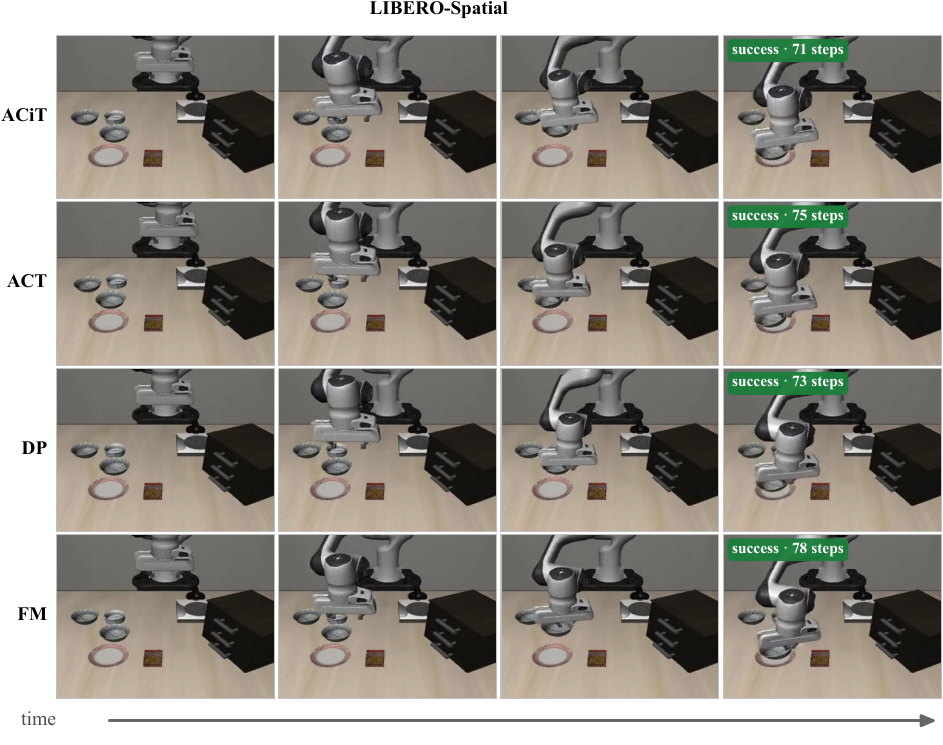}
\caption{Qualitative comparison on LIBERO-Spatial. All four policies solve the episode,
and ACiT finishes in the fewest steps.}
\label{fig:qual-libero}
\end{figure*}

\subsection{Action-Space Dependency Metrics}
\label{app:dependency}

The motivation for the architecture rests on two properties of the action
distribution: its effective dimensionality is far below the number of joints, and
the joints are strongly coupled, with part of that coupling delayed in time. We
measure both on the demonstration data, using only the joint signals
$\mathbf{s}_t,\mathbf{a}_t\in\mathbb{R}^{14}$ and no visual information.

For the action matrix $\mathbf{A}\in\mathbb{R}^{N\times14}$, after centering and
forming $\mathbf{C}=\frac{1}{N-1}\tilde{\mathbf{A}}^{\top}\tilde{\mathbf{A}}$ with
eigenvalues $\lambda_1\ge\dots\ge\lambda_{14}\ge0$, the participation ratio is
\begin{equation}
\mathrm{PR} = \frac{\bigl(\sum_{i=1}^{14}\lambda_i\bigr)^2}{\sum_{i=1}^{14}\lambda_i^2}
\in [1,\,14],
\end{equation}
so that $\mathrm{PR}=1$ means a single effective direction and
$\mathrm{PR}=14$ means fourteen independent ones. Coupling is measured by the
mean pairwise correlation between joints of the left and the right arm, both at
lag zero and maximized over lags in $[-45,45]$ frames.

Table~\ref{tab:dependency} gives the two statistics for the four dual-arm tasks.
Every task sits far below the
ceiling of 14, between 1.44 and 4.42, and the cross-arm correlation is higher at
the best lag than at lag zero on all four tasks. A single instantaneous joint
vector therefore cannot carry either the low effective rank or the delayed
coupling.

\begin{table}[t]
\caption{Action-space statistics on the four dual-arm RoboTwin tasks. Move Can
Pot and Shake Bottle are single-arm and are not listed, because their cross-arm
quantities are undefined.}
\label{tab:dependency}
\begin{center}
\small
\setlength{\tabcolsep}{3.5pt}
\resizebox{\columnwidth}{!}{%
\begin{tabular}{lccccc}
\toprule
Task & PR & within-L & within-R & cross(0) & cross(best) \\
\midrule
Lift Pot & 1.44 & 0.602 & 0.579 & 0.802 & 0.956 \\
Pick Dual Bottles & 2.71 & 0.341 & 0.324 & 0.579 & 0.881 \\
Handover Block & 1.92 & 0.422 & 0.596 & 0.405 & 0.716 \\
Place Dual Shoes & 4.42 & 0.341 & 0.301 & 0.352 & 0.670 \\
\bottomrule
\end{tabular}}
\end{center}
\end{table}

\subsection{Attention Maps and Head Matching}

\begin{figure*}[t!]
\centering
\includegraphics[width=\textwidth]{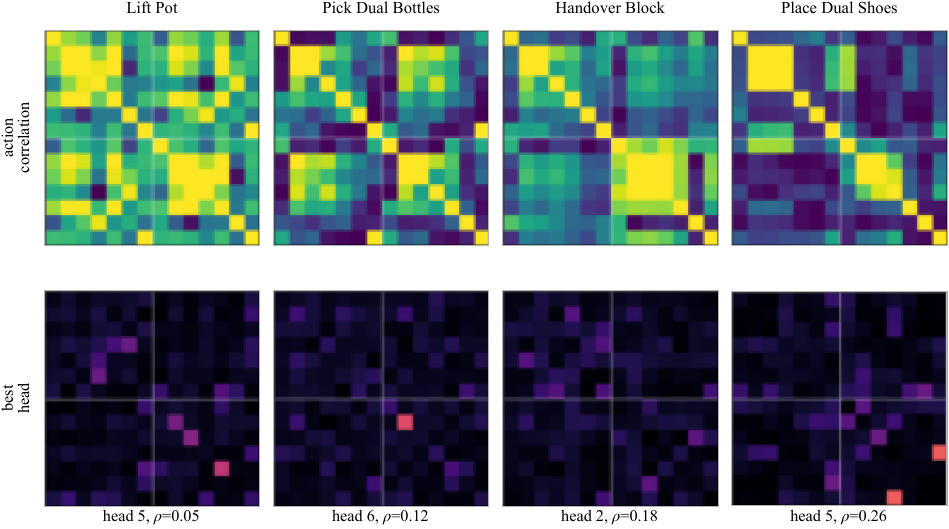}
\caption{Action correlation matrix (left) versus the best-matching
self-attention head (right) for each dual-arm task. The Spearman correlation
over off-diagonal entries is shown above each attention map.}
\label{fig:head-match}
\end{figure*}
\label{app:attention-heads}
Fig.~\ref{fig:head-match} compares the two matrices. For each dual-arm task we
compute the mean 14$\times$14 self-attention map of each head over 5 episodes and
10 frames, symmetrize it, and compare it with the
absolute zero-lag action-correlation matrix using Spearman correlation over
off-diagonal entries. The best head per task is only weakly aligned:
Lift Pot 0.05 (head 5), Pick Dual 0.12 (head 6), Handover 0.18 (head 2), and
Shoes 0.26 (head 5). Thus head-level attention is not a reliable proxy for the
action-correlation structure. We therefore report it as a qualitative
supplement to the quantitative statistics of \S\ref{app:dependency}.

\section{Extended Related Work}
\label{app:related}
\subsection{Proprioceptive History: Representation and Conditioning}
\label{sec:related-proprio}
Action policies are moving from single-frame proprioceptive conditions to
explicit history modeling. In the continuous route, ACT~\citep{act} maps the
current joint state and visual observations to an action chunk with a CVAE.
Diffusion Policy~\citep{dp} conditions on the last $H$ observations,
proprioception included, concatenated into one global vector before conditional
denoising; its own ablation reports that a vision-based policy degrades as $H$ grows, which suggests that a flattened history does not scale. A2A~\citep{a2a} embeds historical proprioceptive sequences into the
latent space of action generation, and Chronos~\citep{chronos} treats observation
history as the latent state of the policy. Meanwhile, another route discretizes
historical actions or state-action trajectories into tokens for autoregressive
VLAs at the cost of losing continuous coupling: VQ-BeT~\citep{vqbet} uses a
residual VQ-VAE to encode continuous action chunks into discrete latent codes
and adds a continuous offset for precision. FAST~\citep{fast} compresses action
chunks in the frequency domain with DCT and BPE, substantially reducing
autoregressive training cost. OmniSAT~\citep{omnisat} further uses B-spline
encoding and multi-stage residual quantization to quantize position, rotation,
and gripper subspaces separately. VITA~\citep{vita} learns a shared discrete
latent space for vision and action and internalizes visual dynamics as an
implicit chain of thought. Both continuous and discrete routes validate the
value of using history~\citep{proprioceptive_transformer,thinkproprio}. However, under either interface, proprioception is still
represented as a flat vector or a one-dimensional token sequence, and the
continuous coupling among joints is not explicitly modeled.

At the fusion level, mainstream methods symmetrically concatenate vision and
proprioception into the same attention stack. A few works instead let the robot
state actively guide vision: RcP~\citep{rcp} uses proprioceptive-state
correspondence to guide robot-centric pooling of image features, extracting
robot-centric visual representations robust to self-distractors.
AVA-VLA~\citep{avavla} uses a recurrent history belief to reweight current visual
tokens. These explorations show that proprioception should not be only a passive
condition, but they stop at perceptual filtering and do not carry structured
joint priors through to action generation.

\subsection{Topology-Aware Morphological Bias}
Standard VLA policies are usually embodiment-agnostic and must infer kinematic
structure from data. GET-Zero~\citep{getzero} proposes a Graph Embodiment
Transformer that uses embodiment graph connectivity as a learned structural bias
in attention, enabling zero-shot adaptation to new joint configurations and link
lengths. Human2LocoMan~\citep{human2locoman} further proposes a Modularized
Cross-embodiment Transformer (MXT) that uses embodiment-specific tokenizers and
detokenizers on a shared Transformer trunk to handle heterogeneous bodies. These
works establish that joints are not independent dimensions and that the kinematic
graph should be an explicit inductive bias. However, existing topology methods
mainly target cross-embodiment transfer and action-interface design. They do not
treat continuous per-joint proprioceptive history as a generative prior, nor let
each joint retrieve its own visual calibration signal. ACiT unifies continuous per-joint history, kinematic topology, and per-joint
visual grounding in a single inverted generation pipeline.

\fi

\end{document}

%% file: math_commands.tex
\usepackage{amsmath,amsfonts,bm}

\def\eqref#1{(\ref{#1})}

\def\1{\bm{1}}

\DeclareMathAlphabet{\mathsfit}{\encodingdefault}{\sfdefault}{m}{sl}
\SetMathAlphabet{\mathsfit}{bold}{\encodingdefault}{\sfdefault}{bx}{n}

